# Deep Multi-modal Fusion of Image and Non-image Data in Disease Diagnosis and Prognosis: A Review

Can Cui[a], Haichun Yang[b], Yaohong Wang[b], Shilin Zhao[c], Zuhayr Asad[a], Lori A. Coburn[d,f], Keith T. Wilson[b,d,f], Bennett A. Landman[e,a], and Yuankai Huo[a,e]

[a]Department of Computer Science, Vanderbilt University, Nashville, TN 37235, USA

[b] Department of Pathology, Microbiology and Immunology, Vanderbilt University Medical Center, Nashville, TN 37215, USA

[c] Department of Biostatistics, Vanderbilt University Medical Center, Nashville, TN 37215, USA

[d] Division of Gastroenterology Hepatology, and Nutrition, Department of Medicine, Vanderbilt University Medical Center, Nashville, TN 37232, USA

[e] Department of Electrical and Computer Engineering, Vanderbilt University, Nashville, TN 37235, USA

[f] Veterans Affairs Tennessee Valley Healthcare System, Nashville, TN 37212, USA

## ABSTRACT

The rapid development of diagnostic technologies in healthcare is leading to higher requirements for physicians to handle and integrate the heterogeneous, yet complementary data that are produced during routine practice. For instance, the personalized diagnosis and treatment planning for a single cancer patient relies on various image (e.g., Radiology, pathology and camera image) and non-image data (e.g., clinical data and genomic data). However, such decision-making procedures can be subjective, qualitative, and have large inter-subject variabilities. With the recent advances in multi-modal deep learning technologies, an increasingly large number of efforts have been devoted to a key question: how do we extract and aggregate multi-modal information to ultimately provide more objective, quantitative computer-aided clinical decision making? This paper reviews the recent studies on dealing with such a question. Briefly, this review will include the (1) overview of current multi-modal learning workflows, (2) summarization of multi-modal fusion methods, (3) discussion of the performance, (4) applications in disease diagnosis and prognosis, and (5) challenges and future directions.

**Keywords:** Multi-modal learning, fusion method, medical data, diagnosis and prognosis

**Highlights:**

- A new "technical-centered" review of deep multi-modal fusion methods (image and non-image data) in disease diagnosis and prognosis is conducted, which is different from the recent "clinical-centered" prior arts.
- Modalities are comprehensively introduced, including image (Radiology, pathology, camera images, etc.) and non-image data (clinical data, lab test, genomics, free text, etc.).
- Existing fusion methods are extensively reviewed and categorized into five families with detailed tables.
- Challenges and future directions of image and non-image medical data fusion are provided.

## 1. INTRODUCTION

Routine clinical visits of a single patient might produce digital data in multiple modalities, including **image data** (i.e., pathology images, Radiology images, and camera images) and **non-image data** (i.e., lab test results and clinical data). The heterogeneous data would provide different views of the same patient to better support various clinical decisions (e.g., disease diagnosis and prognosis [1] [2] [3]). However, such decision-making procedures can be subjective, qualitative, and exhibit large inter-subject variabilities [4] [5]. With the rapid development of artificial intelligence technologies, an increasingly large amount of deep learning-based solutions has been developed for multi-modal learning in medical applications. Deep learning includes high-level abstraction of complex phenomenon within high-dimensional data, which tends to benefit multi-modal fusion in extracting and modeling the complex relationships of different modalities and outcomes [6] [7].

Many works have achieved great success in using a single modality to make a diagnosis or prognosis with deep learning methods [8][9][10]. However, fusing the multi-modal data effectively is not a trivial task in method design because different clinical modalities may contain different information (complementary information of a subject) and have different data formats. Figure 1 summarizes the scope of this review: that the multi-modal data (image and non-image data) from the same patient are utilized for diagnosis or prognosis of diseases. The **image data** can be categorized as Radiology image data, pathology image data, and camera image data. Such imaging data can be further classified as pixel-aligned data (can be spatially registered and overlayed) and pixel-not-aligned data (the pixels in different images do not have spatial correspondence), which might even have different dimensionalities (e.g., 2D, 3D, and 4D). The **non-image data** can be categorized as lab test results such as structured genomic sequences and blood test results, and clinical data including tabular data of demographic features, or free text in the lab test reports. The heterogeneity of such image and non-image data leads to critical challenges in performing multi-modal learning, a family of algorithms in machine learning. For example, 2D pathology images provide micro-level morphology for a tumor while the 3D CT/MRI Radiology images offer macro-level and spatial information of the same tumor. The clinical data and lab test results indicate the molecular, biological, and chemical characteristics, while the structured DNA and mRNA sequences are also involved in clinical decision making. Moreover, image data are typically larger and denser (e.g., millions of pixels), while the non-image data are more sparse with a lower dimensionality. Herein, the heterogeneous formats (e.g., different dimensions, image, free text, and tabular data) require different preprocessing and feature extraction methods, and different types of information require fusion methods that are able to capture the shared and complementary information effectively for rendering better diagnosis and prognosis.

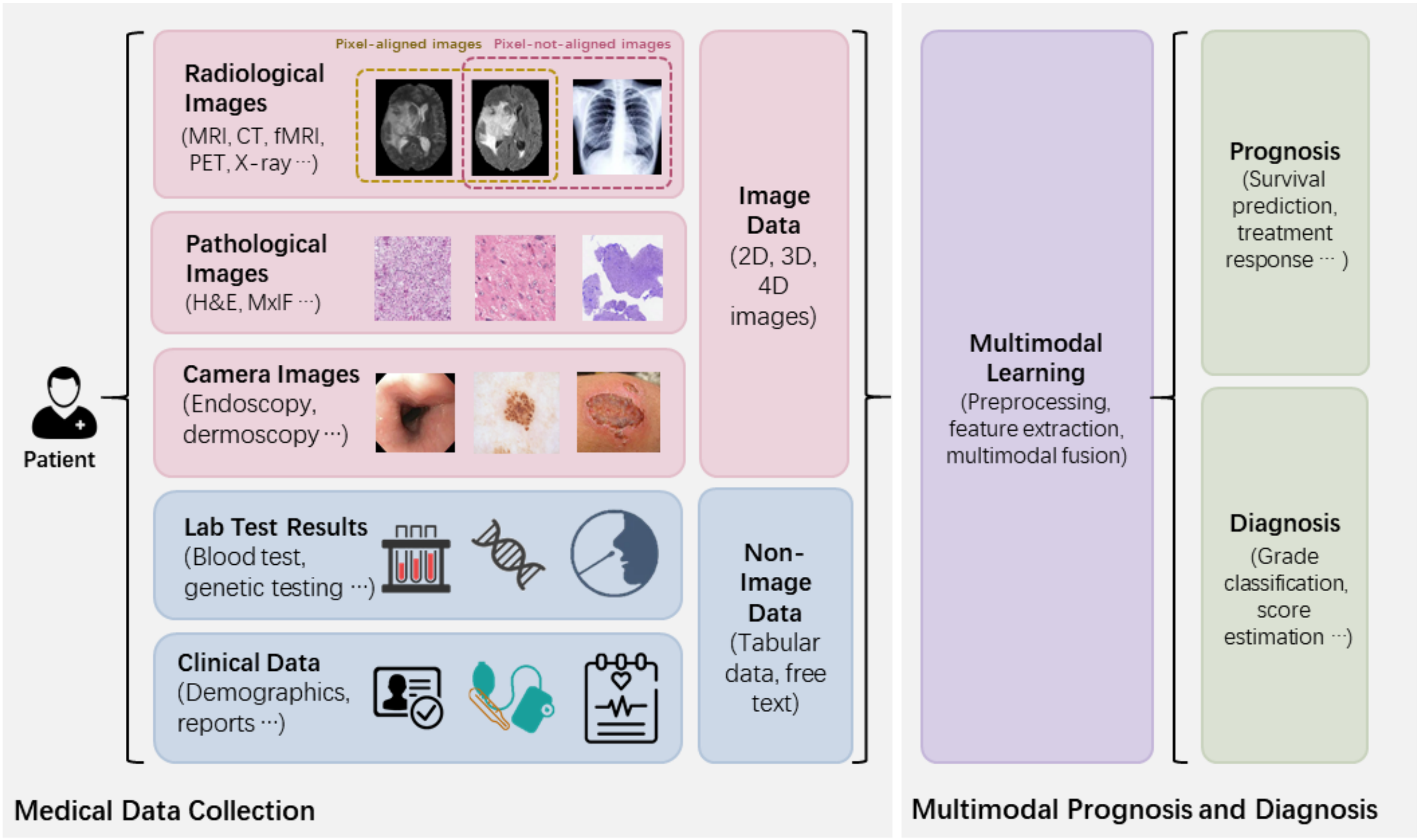

Figure 1. The scope of this review is presented. Multi-modal data containing image data (e.g., Radiology images and pathology images) and non-image data (e.g., genomic data and clinical data) are fused through multi-modal learning methods for diagnosis and prognosis of diseases.

Several surveys have been published for medical multi-modal fusion [11] [12][13][14]. Boehm et al. [11] reviewed the applications, challenges, and future direction of multi-modal fusion in oncology. Huang et al. [12] categorized the fusion methods by the stages of fusion. Schneider et al. [13] and Lu et al. [14] divided the multi-modal learning studies by downstream tasks and modalities. Different from the prior surveys, we review multi-modal fusion techniques from a new perspective of categorizing the methods into operation-based, subspace-based, tensor-based, and graph-based fusion methods. We hope the summary of the fusion techniques can foster new methods in medical multi-modal learning.

In this survey, we collected and reviewed 34 related works published within the last 5 years. All of them used deep learning methods to fuse image and non-image medical data for prognosis, diagnosis, or treatment prediction. This survey is organized in the following structure: §2 provides an overview of multi-modal learning for medical diagnosis and prognosis; §3 briefly introduces the data preprocessing and feature extraction for different uni-modalities, which is the prerequisite for multi-modal fusion; §4 summarizes categorized multi-modal fusion methods, and their motivation, performance, and limitations are discussed; §5 provides a comprehensive discussion and future directions; and §6 is a conclusion.

# 2. OVERVIEW

## 2.1 Study selection

This survey only includes published studies (with peer-review) that fuse both image and non-image data to make a disease diagnosis or prognosis in the past five years. All of them used feature-level deep learning-based fusion methods for multi-modal data. A total of 34 studies that satisfied these criteria are reviewed in this survey.

## 2.2 Workflow

A generalized workflow of collected studies is shown in Figure 2. Typically, data preprocessing, uni-modal feature extraction, multi-modal fusion, and predictor sections are included in the workflow. Due to the heterogeneity of image and non-image modalities, it is unusual to fuse the original data directly. Different modalities always have separate methods of data preprocessing and feature extraction. For multi-modal learning, fusion is a crucial step, following the uni-modal data preprocessing and feature extraction steps that are the prerequisites. §3 and §4 will introduce and discuss the uni-modal feature preparation and multi-modal fusion separately.

Based on the type of inputs for multi-modal fusion, the fusion strategies can be divided into feature-level fusion and decision-level fusion [12]. **Feature-level fusion** contains early fusion and intermediate fusion. For **decision-level fusion**, which is also referred to as late fusion, the prediction results of uni-modal models (e.g., probability logits or categorical results from uni-modal paths in classification tasks) are fused for multi-modal prediction by majority vote, weighted sum, or averaging, etc. The fusion operation is relatively simple and there is no need to retrain the uni-modal models at the fusion stage. As for feature-level fusion, either the extracted high-dimensional features or the original structured data can be used as the inputs. Compared with decision-level fusion, feature-level fusion has the advantage of incorporating the complementary and correlated relationships of the low-level and high-level features of different modalities [12] [15], which leads to more variants of fusion techniques. This survey mainly focuses on categorizing the methods of feature-level fusion, but also compares with the decision-level fusion.

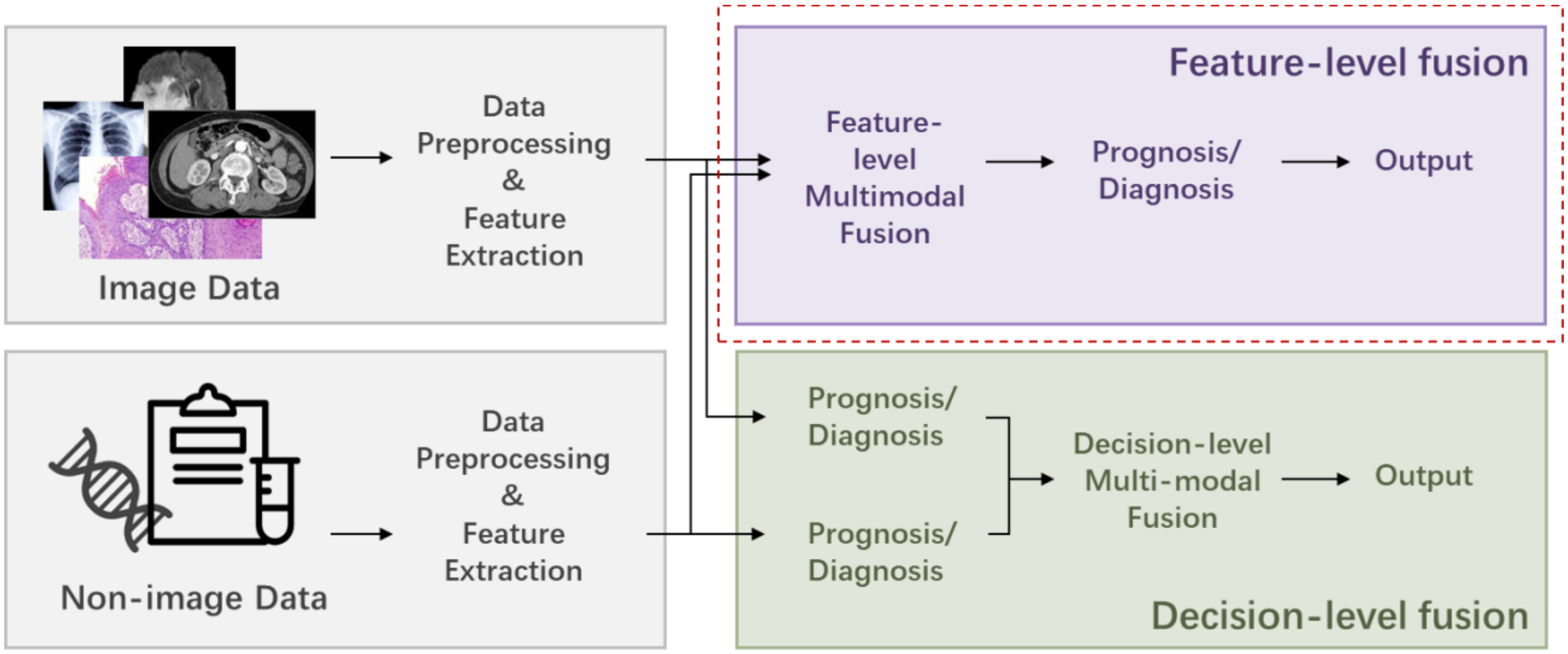

Figure 2. The overview of the multi-modal learning workflow is presented. Due to the heterogeneity of different modalities, separate preprocessing

methods and feature extraction methods are used for each modality. For feature-level fusion, the extracted features from uni-modals are fused. Note that the feature extraction methods can be omitted, because some data can be fused directly (such as the tabular clinical features). As for decision-level fusion, the different modalities are fused in the probability or final predication level. Because feature-level fusion contains more variants of fusion strategies, we mainly focus on reviewing the feature-level fusion methods, but also compares with the decision-level fusion.

### 2.3 Diagnosis and prognosis tasks and evaluation metrics

In this survey, multi-modal fusion is applied to the disease diagnosis and prognosis. The disease diagnosis tasks include classifications such as disease severity, benign or malignant tumors, and regression of clinical scores. Prognosis tasks include survival prediction and treatment response prediction.

After obtaining the multi-modal representations, multi-layer perceptrons (MLP) were used by most of the studies to generate the prognosis or diagnosis results. The specific tasks of diagnosis and prognosis can be categorized into regression or classification tasks based on discrete or continuous outputs. To supervise the modal training, the cross-entropy loss was usually used for classification tasks, while the mean square error (MSE) was a popular choice for regression tasks. To evaluate the results, the area under the curve (AUC), receiver operating characteristics curve (ROC), accuracy, F1-score, sensitivity, and specificity metrics were commonly used for classification, while the MSE was typically used for regression. However, although the survival prediction was treated as a time regression task or a classification task of long-term/short-term survival, the Cox proportional hazards loss function [16] was popular in survival prediction tasks. To evaluate the survival prediction models, the concordance index (c-index) was widely used to measure the concordance between the predicted survival risk and real survival time.

## 3. UNI-MODAL DATA PREPROCESSING AND FEATURE EXTRACTION

Due to multi-modal heterogeneity, separate preprocessing and feature extraction methods/networks are required for different modalities to prepare uni-modal features for fusion. As shown in Table 1, our reviewed studies contain image modalities such as pathology images (H&E), Radiology images (CT, MRI, X-ray, fMRI), and camera images (clinical images, macroscopic images, dermatoscopic images); and non-image modalities such as lab test results (genomic sequences) and clinical features (free-text reports and demographic data). In this section, we briefly introduce these data modalities and summarize the corresponding data preprocessing and feature extraction methods.

### 3.1 Image data

*Pathology images*

Pathology images analyze cells and tissues at a microscopic level, which is recognized as the “gold standard” for cancer diagnosis [17]. The 2D Hematoxylin-Eosin (H&E) stained pathology images with 3 channels is a widely used one. Yet, the whole slide image (WSI) of pathology images usually cannot be processed directly because of its gigantic size. So, the smaller patches are always cropped from the informative regions of interests (ROIs) of WSIs to fit the computation memory. To define the ROIs, some reviewed studies used the diagnostic ROIs manually annotated by experts [18] [19] or predicted by the pre-trained segmentation models [20], [21], while some reviewed studies instead selected ROIs from the representative region [22], foreground [23] or dense region [24], [25] based on pixel intensity. Also, the color space can be converted from RGB to HSV for H&E images for higher intensity contrast, and always standardized [26] [27] or calibrated to standard images [16], [20] for data consistency. To generate features from cropped image patches, 2D conventional neural networks (CNN) are popular options for learning-based features. Moreover, finetuning pre-trained by natural images (e.g., ImageNet) is usually preferred especially for small datasets [21] [53] [27] [22], though there is a concern for the large gap between natural images and pathology images [28]. Except for the learning-based methods, some other works [29], [30] used conventional feature extraction methods (e.g., CellProfiler [31]) to extract statistical and structured features about shapes, texture, intensity, etc. Meanwhile, multiscale is a common mechanism to extract complementary and shared information from different level of pathology images [18] [32] [33] [21] [18]. As for the supervision, if only the WSI-level label is provided for patches (e.g., the patches from ROIs extracted based on intensity), multi-instance learning (MIL) is always applied to aggregate the information of patches in bags for supervision [34], [35].

*Radiology images*

Radiology imaging supports medical decisions by providing visible image contrasts inside the human body with radiant energy, including magnetic resonance imaging (MRI), computed tomography (CT), positron emission tomography (PET) and X-ray, etc. To embed the intensity standardized 2D or 3D radiology images into feature representations with learning-based encoders [36], [37] [38] [39] [15] [40] or conventional radiomics methods [36], [37] [40] or both [37] [36], skull-stripping [41], affine registration [41], foreground extraction [42], lesion segmentation [41] [18] [41] [36] [37] were used correspondingly in some reviewed works to define the ROIs at first. And then, the images were resized or cropped to a smaller size for feature extraction. In order to further reduce the image dimension to fit computation memory but keep essential information, some works used 2D maximum intensity projection (MIP) [43], and some works used the representative slices with maximum tumor diameters [18] [37] to convert the 3D volume to 2D. Functional MRI (fMRI) is another radiological modality used in some reviewed studies that investigated autism spectrum disorder (ASD) and Alzheimer's disease (AD) [44], [45],. The images of brains were divided into multiple regions by the template. Then, the Pearson correlation coefficient between two brain regions was calculated to form the functional connectivity matrix. The matrix was finally vectorized for classification.

*Camera Images*

In addition to pathology and radiology images, some other kinds of medical images captured by optical color cameras are categorized as camera images; examples of these camera images include dermoscopic and clinical images for skin lesions [46] [47], endoscopic images to examine the interior of a hollow organ or cavity of the body [48], and the funduscopic images photographing the rear of eyes [49]. Different from pathology images, camera images can be taken directly while the sectioned and stained sample slides are not required. Also, most of the camera images are 3-channel color images but in a smaller size than pathology images. 2D pre-trained CNN networks [50] or pre-trained transformer models [51] are usually applied to the whole images or detected lesions.

## 3.2 Non-image data

The non-image modalities contain lab test results and clinical features. Laboratory tests check a sample of blood, urine, or body tissues, access the cognition and psychological status of patients, and analyze genomic sequences, etc. clinical features include demographic information and clinical reports. These modalities are also essential to diagnosis and prognosis in clinical practice. The non-image data on the reviewed works can be briefly divided into structured data and free-text data for different preprocessing and feature extraction methods.

*Structured data*

Most of the clinical data and lab test results in the reviewed works are structured data and can be converted to feature vectors easily. In preprocessing, categorical clinical features were usually converted through one-hot encoding, while the numerical features were standardized [43] [36] [37]. Specially, Cai et al. [51] used soft one-hot encoding by setting the elements that are as 0 in the standard one-hot encoding as 0.1 to make contributions to the propagation. As the genomic data are in high dimension, some feature selection methods such as the highest variance were used to extract the most expressive features [22] [26] [52]. The missing value is a common problem for some structured data. The ones with a high missing rate were usually discarded directly, while the other missing data were imputed with the average value, mode value, or values of similar samples selected by K-nearest neighbors (KNN) [40] [15], and some works added missing status as features [53] [54].

*Free-text data*

Clinical reports capture clinicians' impressions of diseases in the form of unstructured text. In order to deal with the free text data and extract informative features from the free-text, natural language processing techniques are implemented. For example, Chauhan et al. [55] prepared the tokenization of the text extracted by ScispaCy [56]. Then, the BERT [57] model initialized by weights pre-trained on the scientific text [58] was used to embed the tokenization. Furthermore, the language model trained by medical data such as ClinicalBERT [59] was tried in the work [60] for text embeddings and compared its performance with BERT in multi-modal prediction.

After using the above modal-specific preprocessing and feature extraction methods, the uni-modal representations could be converted to feature maps or feature vectors. For feature vectors, in order to learn more expressive features with expected dimensions, Sappagh et al. [40] used principal component analysis (PCA), to reduce the dimension of radiomics features. Parisot et al. [44] explored different feature reduction methods such as recursive feature elimination (RFE), principal component analysis (PCA), and the autoencoder to reduce the feature dimension of the vectorized functional connectivity matrix. In contrast, Yan et al. [28] used the denoising autoencoder to enlarge the dimension of low-dimensional clinical features to avoid being overwhelmed in feature-level fusion by high-dimensional image features. For similar purpose, Yoo et al. [41] replicated and scaled the clinical features, while Cui et al. [53] deconvoluted the feature vectors to the same size as feature maps of image modalities. In terms of aggregating multiple feature vectors, Sappagh et al. [40] used bidirectional long short-term memory (biLSTM) models to handle the vectorized time-series features, and Lu et al. [23] applied attention-based multi-instance learning to accumulate the feature vectors in bags to bag-level representations. As for the image feature maps learned by CNNs, these feature maps could be used for fusion directly in order to keep the spatial information [53] [61], vectorized with pooling layers [37] [15] [62] or split by pixel/voxels as the tokens to feed transformers [60].

Unimodal feature extraction can be unsupervised or supervised. Note that if the unimodal features can be trained before fusion, the maximum number of available samples for each modality can be used for better uni-modal model performance and hopefully better unimodal features [26] [37] [63]. Regarding the relationship of unimodal feature extraction with the fusion, the uni-modal feature extraction section can be independent to the fusion section [39] [37], which is known as early-level fusion, or can be trained from scratch or finetuned with the fusion section end-to-end [32] [18] as the intermediate-level fusion. Note that if the unimodal features can be trained before fusion, the maximum number of available samples for each modality can be used for better uni-modal models and hopefully the corresponding unimodal features [63].

## 4. MULTI-MODAL FUSION METHODS

Fusing the heterogeneous information from multi-modal data to effectively boost prediction performance is a key pursuit and challenge in multi-modal learning [64]. Based on the type of inputs for multi-modal fusion, the fusion strategies can be divided into feature-level fusion and decision-level fusion [11]. **Decision-level fusion** integrates the probability or categorical predictions from uni-modal models using simple operations such as averaging, weighted vote, majority vote, or a meta classifier with trainable layers for unimodal probability [15][65][47][39][66], to make a final multi-modal prediction. For the decision-level fusion, the prediction of unimodality can be learned separately and be independent to the fusion stage. It can fuse any combination of multi-modalities without further adjustment in the testing phase. So, it may be preferable for flexibility and simplicity. and it can tolerate the missing modality situation. Sometimes the decision-level fusion achieved better performance than the feature-level fusion. For example, Wang et al. [66] implemented a learnable weighted sum mechanism based on unimodal uncertainty to fuse the prediction of different modalities, which outperformed the intermediate feature-level fusion. Huang et al. [65] showed that decision-level fusion also outperformed feature-level fusion in their experiments of pulmonary embolism detection. However, the decision-level fusion may lack the interaction of the features. For the modalities with dependent or correlated features, feature-level fusion might be more preferable. Some other works [67] [15] also showed that their proposed feature-level integration performed better than decision-level fusion. On the other hand, **feature-level fusion** fuses the original data or extracted features of heterogeneous multi-modals into a compact and informative multi-modal hidden representation to make a final prediction. Compared with decision-level fusion, more variants of feature-level fusion methods have been proposed to capture the complicated relationship of features from different modalities. This survey reviews these methods and categorizes them into operation-based, subspace-based, attention-based, tensor-based, and graph-based methods. The representative structures of these fusion methods are displayed in Figure 3, and the fusion methods of reviewed studies are summarized in Table 1.

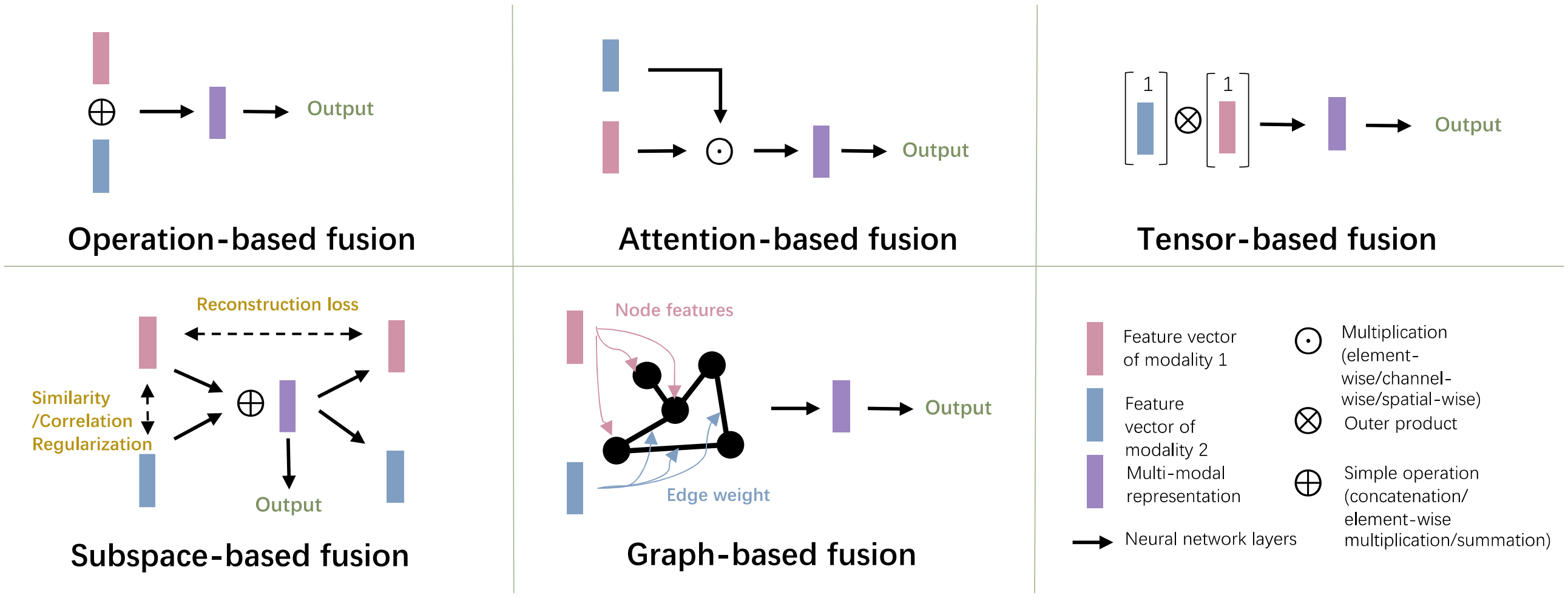

Figure 3. Representative structures of different feature-level multi-modal fusion methods.

### 4.1 Operation-based fusion methods

To combine different feature vectors, the common practice is to perform simple operations of concatenation, element-wise summation, and element-wise multiplication. These practices are parameter-free and flexible to use, but the element-wise summation and multiplication methods always require the feature vectors of different modalities to be converted into the same shape. Many early works used one of the simple operations to show that multi-modal learning models outperforms uni-modal models [40] [23] [15] [19] [47] [46] [39] [41]. Although the operation-based fusion methods are simple and effective, they might not exploit the complex correlation between heterogeneous modalities. Also, the long feature vectors generated by the concatenation may lead to overfitting when the amount of training data is not sufficient [68][69]. More recently, Holste et al. [15] compared these three operation-based methods in the task of using clinical data and MRI images for breast cancer classification. The low-dimensional non-image features were processed by fully connected layers (FCN) to the same dimension of image features before fusion. The results showed that the three operations performed comparably ($p$-value > 0.05), while the element-wise summation and multiplication methods required less trainable parameters in the following fully connected layers. After comparing the learned non-image features by FCN and the original non-image features, the former ones achieved superior performance. Meanwhile, the concatenation of the feature vectors outperformed the concatenation of logits from the uni-modal data. Yan et al. [33] investigated the influence of the dimension of uni-modal features on the uni-modal performance using concatenation fusion. They hypothesized that the high-dimensional vectors of image data would overwhelm the low-dimension clinical data. To keep the rich information of the high-dimensional features for a sufficient fusion, they used the denoising autoencoder to increase the dimension of clinical features. Zhou et al. [63] proposed a three-stage uni-modal feature learning and multi-modal feature concatenation pipeline, where every two modalities were fused at the second stage and all three modalities were fused at the third stage, in order to use the maximum number of available samples when some modalities were missing.

### 4.2 Subspace-based fusion methods

The subspace methods aim to learn an informative common subspace of multi-modality. A popular strategy is to enhance the correlation or similarity of features from different modalities. Yao et al. [29] proposed a DeepCorrSurv model and evaluated the survival prediction task. Inspired by the conventional canonical correlation analysis (CCA) method [53], they proposed an additional CCA-based loss for the supervised FCN network to learn the more correlated feature space of features from two modalities. The proposed methods outperformed the conventional CCA methods by learning the non-linear features and the supervised correlated space. Zhou et al. [42] designed two similarity losses to enforce the learning of modality-shared information. Specifically, a cosine similarity loss was used to supervise the features learned from these two modalities, and a loss of hetero-centre distance was designed to penalize the distance between the center of clinical features and CT features belonging to each class. In their experiments, the accuracy dropped from 96.36 to 93.18 without these similarity losses. Li et al. [70] used the average of L1-norm and L2-norm loss to improve the similarity of the learned uni-modal features from pathology images and genes before concatenating them as a multi-modal representation. The learned similar features can then be fused by concatenation as the multi-modal representation.

Another study fused the feature vectors from 4 modalities with the subspace idea in the diagnosis task of 20 cancer types [52]. Inspired by the SimSiam network [71], they forced the feature vectors from the same subject to be similar by a margin-based hinge-loss. Briefly, cosine similarity scores between the uni-modal features from the same patient were maximized, whereas the ones from different patients were minimized. The feature similarities of different patients were only penalized within a margin of the feature similarity. Such a regularity enforced similar feature representation from the same patient, while avoiding mode collapse.

Another strategy in the subspace-based fusion method is to learn a complete representation subspace with the encoder-decoder structures. Ghosal et al. [72] decoded the mean vectors of multi-modal features and used the reconstruction loss to force the mean vectors to contain the complete information of different views. The mean vectors with additional decoder and reconstruction loss achieved superior classification accuracy as compared with the counterparts without such loss functions. Similarly, Cui et al. [37] also used the autoencoder backbone to learn the complete representation, but some modalities were randomly dropout and reconstructed by the mean vector generated from the available modalities to improve prediction accuracy and be more robust to the testing data with missing modalities.

## 4.3 Attention-based fusion methods

Attention-based methods computed and incorporated the importance scores (attention weights) of multi-modality features when performing aggregation. This progress simulated routine clinical practice. For example, the information from clinical reports of a patient may inform the clinicians to pay more attention to a certain region in an MRI image. Duanmu et al. [61] built an FCN path for non-image data along with a CNN path for image data. The learned feature vectors from the FCN path were employed as the channel-wise attention for the CNN path at the corresponding layers. The low-level and high-level features of different modalities can be fused correspondingly, which achieved a better prediction accuracy than simple concatenation. Schulz et al. [18] concatenated the learned feature vectors from three modalities by an attention layer, which weighted the importance for the downstream task. Chen et al. [27] calculated the co-attention weight to generate the genomic-guided WSI embeddings. Similarly, Lu et al. [73] proposed a symmetric cross attention to fuse the genomic data and pathology image embeddings of glioma tumors for multitask learning, while Cai et al. [51] proposed an asymmetrical multi-head cross attention to fuse the camera images and metadata for skin classification. Li et al. [32] aggregated pathology and clinical features to predict the lymph node metastasis (LNM) of breast cancer. To utilize the Gigapixel WSIs, they proposed the multi-modal attention-based MIL to achieve patient-level image representation. The clinical features were further integrated to form instance-level image attention for the downstream task. The experiments showed that the proposed attention-based methods outperformed both the gating-based attention used by Chen et al. [26] and a bag-concept layer concatenation [74]. Guan et al. [38] applied the self-attention mechanism [21] in their concatenated multi-modal feature maps. They tiled and transformed the clinical feature vectors to the same shape of the image feature matrix to keep the spatial information in an image feature map. Their performance surpassed both the concatenation and another subspace method using a similarity loss [55]. In addition to the MLP and CNN, the attention mechanism was also applied to the graph model for multi-modal learning in the medical domain. Cui et al. [53] built a graph where each node was composed of image features and clinical features with category-wise attention. The influence weights of neighbouring nodes were learned by the convolution graph attention network (con-GAT) and novel correlation-based graph attention network (cor-GAT). The attention value was used to update the node features for the final prediction. More recently, the transformer models were widely used in multi-modal learning [76], and they were adapted to medical field. Jacenkow et al. [77] exploited the unimodally pre-trained transformer-based language model BERT [57] and finetuned it after adding the image tokens for multi-modal learning. Li et al. [60] used the radiology images and radiology free-text report to finetune the visual-text transformer models pre-trained by general image-language pairs not specific for medical domains. Meanwhile, different visual-text backbone, unimodal pre-trained models and training strategy were compared in their work.

The above attention-based fusion methods rescaled features through complementary information from another modality, while Pölsterl et al. [54] proposed a dynamic affine transform module that shifted the feature map. The proposed modules dynamically produced scale factor and offset conditional on both image and clinical data. In such a design, the affine transform was added ahead of the convolutional layer in the last residual block to rescale and shift the image feature maps. As a result, the high-level image features can interact with the compacted clinical features, which outperformed the simple concatenation and channel-wise attention-based methods [61].

## 4.4 Tensor-based fusion methods

The tensor-based fusion methods conducted outer products across multi-modality feature vectors to form a higher order co-occurrence matrix. The high-order interactions tend to provide more predictive information beyond what those features can provide individually. For example, blood pressure rising is common when a person is doing high-pressure work, but it is dangerous if there are also symptoms of myocardial infarction and hyperlipidemia [42]. Chen et al. [26] proposed pathomic fusion to make prognosis and diagnosis utilizing pathology images, cell graphs, and genomic data. They used the tensor fusion network with a Kronecker product [78] to combine the uni-modal, bimodal, and trimodal features. To further control the expressiveness of each modality, a gated-attention layer [79] was added. Wang et al. [80] not only used the outer product for inter-modal feature interactions, but also for intra-modal feature interactions. It surpassed the performance of the CCA-based method known as DeepCorrSurv [29]. More recently, Braman et al. followed the work of pathomic fusion [26] and extended it from three modalities to four modalities. Also, an additional orthogonal loss was added to force the learned features of different modalities to be orthogonal to each other, which helped to improve feature diversity and reduce feature redundancy. They showed that their methods outperformed the simple concatenation and the original Kronecker product.

## 4.5 Graph-based fusion method

A graph is a non-grid structure to catch the interactions between individual elements represented as nodes. For disease diagnosis and prognosis, nodes can represent the patients, while the graph edges contain the associations between these patients. Different from CNN-based representation, the constructed population graph updates the features for each patient by aggregating the features from the neighbouring patients with similar features. To utilize complementary information in the non-imaging features, Parisot et al. [44] proposed to build the graph with both image and non-image features to predict autism spectrum disorder and Alzheimer's disease. The nodes of the graph were composed of image features extracted from fMRI images, while the edges of the graph were determined by the pairwise similarities of image (fMRI) and non-image features (age, gender, site, and gene data) between different patients. Specifically, the adjacency matrix was defined by the correlation distance between the subject's fMRI features multipled with the similarity measure of non-image features. Their experiment showed that the proposed GCN model outperformed MLP of multi-modal concatenation. Following this study, Cao et al. [45] built graphs similarly but proposed to use the edge dropout and DeepGCN structure with residual connection instead of the original GCN for deeper networks and thus avoid overfitting, which achieved better results.

Table 1: Overview of image and non-image modalities, number of subjects, tasks multi-modal fusion methods and performance comparison of reviewed studies.

| | Study | Modalities | Subjects | Tasks | Fusion Strategy | Fusion details | Performance comparison (uni-/multi-modal) * | Performance comparison (different fusion methods) |
|---|---|---|---|---|---|---|---|---|
| 1 | Holste et al. 2021 [15] | MRI images, clinical features | 17,046 samples of 5,248 subjects | Classification of breast cancer | Operation | Element-wise multiplication/element-wise summation/ concatenation of learned uni-modal features or direct features. | [AUC] Images: 0.860, clinical features: 0.806, All: 0.903 (p-value < 0.05) | [AUC] Concatenation: 0.903, sum: 0.902, multiplication: 0.896, probability fusion: 0.888 (p-value >0.05) |
| 2 | Lu et al. 2021 [23] | H&E images, clinical features | 29,107 patients. External testing set: 682 patients. (Partially public) | Classification of primary or metastatic tumor and origin site. | Operation | Concatenation of clinical features and the learned pathology image feature | [Top-1 accuracy] Image: about 0.74, image+sex: about 0.808, image+sex+site: about 0.762 (metastatic tumors) | - |
| 3 | Sappagh et al. 2020 [40] | MRI and PET images Neuropsychology data, cognitive scores, assessment data | 1,536 patients from ADNI. (Public) | Classification of AD and prodromal status. Regression of 4 cognitive scores | Operation | Concatenation of learned static features and learned time-series uni-modal features from 5 stacked CNN-biLSTM. | [Accuracy] Five modalities: 92.62, four modalities: 90.45, three modalities: 89.40, two modalities: 89.09. (Regression performance is consistent with the classification) | - |
| 4 | Yan et al. 2021 [33] | Pathology images, clinical features | 3,764 samples of 153 patients. (Public) | Classification of breast cancer | Operation | Concatenation of increased-dimensional clinical features and multi-scale image features | [Accuracy] Image+clinical features: 87.9, clinical features: 78.5, images: 83.6 | - |
| 5 | Mobadersany et al. 2018 | H&E images, genomic data | 1,061 samples of 769 patients from the TCGA-GBM and | Survival prediction of gliomas tumor | Operation | Concatenation of genomic biomarkers and learned pathology image features | [C-Index] Images+gene: 0.774, image: 0.745, gene: 0.746. (p-value < 0.05) | - |
| | [19] | | TCGA-LGG. (Public) | | | | | |
| 6 | Yolland et al. 2018 [46] | Macroscopic images, dermatoscopic images, clinical features | 2,917 samples from ISIC. (Public) | Classification of skin lesion | Operation | Concatenation of clinical features and learned image features. | [AUC] Dsc + macro + clinical: 0.888, dsc + macro: 0.888, macro: 0.854, dsc: 0.871 | - |
| 7 | Silva et al. 2020 [22] | Pathology images, mRNA, miRNA, DNA, CNV clinical features | 11,081 patients of 33 cancer types from TCGA. (Public) | Pancancer survival prediction | Operation Attention | Attention weighted element-wise summation of uni-modal features | [C-Index] Clinical: 0.742, mRNA: 0.763, miRNA: 0.717, DNA: 0.761, CNV: 0.640, WSI:0.562,Clinical+mRNA+DNAm: 0.779, all six modalities: 0.768 | - |
| 8 | Kawahara et al. 2019 [47] | Clinical images, dermoscopic images, clinical features | 1,011 cases. (Public) | Classification of skin lesion | Operation | Concatenation of learned uni-modal features. | [Accuracy] Clinical images + clinical features: 65.3 dermoscopic images + clinical features: 72.9 all modalities: 73.7 | - |
| 9 | Yoo et al. 2019 [41] | MRI images, clinical features | 140 patients | Classification of brain lesion conversion | Operation | Concatenation of learned images features and the replicated and rescaled clinical features | [AUC] Images: 71.8, images + clinical: 74.6 | |
| 10 | Yao et al. 2017 [29] | Pathology images , genomic data | 106 patients from TCGA-LUSC, and 126 patients from the TCGA-GBM. (Public) | Survival prediction of lung cancer and brain cancer | Operation Subspace | Maximum correlated representation supervised by the CCA-based loss | [C-index] Pathology images: 0.5540, molecular: 0.5989, images + molecular: 0.6287. (LUSC). Similar results on other two datasets. | [C-index] Proposed: 0.6287, SCCA: 0.5518, DeepCorr+DeepSurv [16]: 0.5760 (LUSC). Similar results on other two datasets. |
| 11 | Cheerla et al. 2019 [52] | Pathology images, Genomic data (gene expression, microRNA, clinical features) | 11,160 patients from TCGA (Nearly 43% of patients miss modalities). (Public) | Survival prediction of 20 types of cancer. | Operation Subspace | The average of learned uni-modal features, while a margin-based hinge-loss was used to regularize the similarity of learned uni-modal features. | [C-index] Clinical+miRNA+mRNA+pathology: 0.78, clinical+miRNA: 0.78, clinical+mRNA: 0.60, clin+miRNA+mRNA:0.78, clinical+miRNA+pathology :0.78 | - |
| 12 | Li et al. 2020 [70] | Pathology images genomic data | 826 cases from the TCGA-BRCA. (Public) | Survival prediction of breast cancer | Operation Subspace | Concatenation of the learned uni-modal features that was regularized by a similarity loss. | [C-index] Images+gene: 0.7571, gene: 0.6912, image: 0.6781. (p-value < 0.05) | - |
| 13 | Zhou et al. 2021 [42] | CT images, laboratory indicators, clinical features | 733 patients | Classification of COVID-19 severity | Operation Subspace | Concatenation of the learned uni-modal features that was regularized by a similarity loss. | [Accuracy] Clinical features: 90.45, CT+clinical features: 96.36 | [Accuracy] Proposed: 96.36, proposed wo/ similarity loss: 93.18 |
| 14 | Ghosal et al. 2021 [72] | Two fMRI paradigms images, genomic data (SNP) | 1) 210 patients from the LIBD institute. 2) External testing set: 97 patients from BARI institute. | Classification of neuropsychiatric disorders | Operation Subspace | Mean vector of learned uni-modal features, supervised by the reconstruction loss. | - | [AUC] Proposed: 0.68, auto-encoder: 0.62, encoder only: 0.59 (LIBD) The external test set showed the same trend of results. |
| 15 | Cui et al. 2022 [37] | H&E and MRI images, DNA, demographic features | 962 patients (170 with complete modalities) from TCGA-GBMLGG and BraTs (Public) | Survival prediction of glioma tumors | Operation Subspace | Mean vector of learned uni-modal features with modality dropout, supervised by the reconstruction loss. | [C-index] Pathology: 0.7319 radiology: 0.7062, DNA: 0.7174, demographics: 0.7050, all: 0.7857 | [C-index] Proposed: 0.8053, pathomic fusion [81]: 0.7697, deep orthogonal[36]: 0.7624 |
| 16 | Schulz et al. 2021 [18] | CT, MRI and H&E images, genomic data | 1) 230 patients from the TCGA-KIRC. 2) External testing set: 18 patients. (Public) | Survival prediction of clear-cell renal cell carcinoma | Operation Attention | Concatenation of learned uni-modal features with an attention layer | [C-index] Radiology: 0.7074, pathology: 0.7424, rad+path: 0.7791. (p-value < 0.05). external test set showed similar results | - |
| 17 | Cui et al. 2021 [53] | CT images, clinical features | 924 cases of 397 patients | Lymph node metastasis prediction of cell carcinoma | Operation Attention | Concatenation of learned uni-modal features with category-wise contextual attention as graph node attributes. | - | [AUC] Proposed: 0.823, logestic regression: 0.713, attention gated [82]: 0.6390, deep insight [83]: 0.739 |
| 18 | Li et al. 2021 [32] | H&E images, clinical features | 3,990 cases | Lymph node metastasis prediction of breast cancer | Operation Attention | Attention-based MIL for WSI-level representation, whose attention coefficients | [AUC] Clinical: 0.8312, image: 0.7111 | [AUC] Proposed: 0.8844, concatenation:0.8420 , gating attention |
| | | | | | | were learned from both modalities | | [26]: 0.8570, M3DN [74]: 0.8117 |
| 19 | Duanmu et al. 2020 [61] | MRI images, genomic data, demographic features | 112 patients | Response prediction to neoadjuvant chemotherapy in breast cancer | Operation Attention | The learned feature vector of non-image modality was multiplied in channel-wise way with the image features at multiple layers. | [AUC] Image: 0.5758 image and non-image: 0.8035 | [AUC] Proposed: 0.8035, concatenation: 0.5871 |
| 20 | Guan et al. 2021 [38] | CT images, clinical features | 553 patients | Classification of esophageal fistula risk | Operation Attention | Self-attention on the concatenation of learned uni-modal features. Concatenation of all paths in the end. | [AUC] Images: 0.7341 clinical features [84]: 0.8196 | [AUC] Proposed: 0.9119, Concate: 0.8953, Ye et al. [85]: 0.7736, Chauhan et al. [55]: 0.6885, Yap et al. [46] :0.8123 |
| 21 | Pölsterl et al. 2021 [54] | MRI images, clinical features | 1,341 patients for diagnosis and 755 patients for prognosis. (Public) | Survival prediction and diagnosis of AD | Operation Attention | Dynamic affine transform module | [C-index] Images: 0.599, images+clinical: 0.748 | [C-index] Proposed: 0.748, FiLM [86]: 0.7012, Duanmu et al. [61]: 0.706, concatenation: 0.729 |
| 22 | Wang et al. 2018 [87] | X-ray images, free-text radiology reports | 1) Chest X-ray 14 dataset, 900 cases from a hand-labeled dataset. 2) 3,643 cases from the OpenI. (Partially public) | Classification of thorax disease | Operation Attention | Learned image features and word embedding-fed LSTM models and generated multi-level attention weights for image and text. | [Weighted accuracy] Text reports: 0.978, images: 0.722, images + text reports: 0.922. (Chest X-rays14). Similar results on other two datasets. | - |
| 23 | Chen et al. 2020 [26] | H&E images, DNA and mRNA | 1) 1,505 cases from 769 patients from TCGA-GBM/LGG. 2) 1,251 cases of 417 patients from TCGA-KIRC. (Public) | Survival prediction and grade classification of glioma tumors and renal cell carcinoma | Operation Attention Tensor Fusion | Kronecker product of different modalities. And a gated-attention layer was used to regularize the unimportant features | [C-index] Images (CNN): 0.792, images (GCN): 0.746, gene: 0.808,images+gene: 0.826. (GBM/LGG) Similar results on the other dataset. | [C-index]: Proposed: 0.826, Mobadersany et al. [19]: 0.781. (p-value<0.05) (GBM/LGG) Similar results on the other dataset. |
| 24 | Wang et al. 2020 [80] | Pathology images, genomic data | 345 patients from TCGA (Public) | Survival prediction of breast cancer | Operation Tensor Fusion | Inter-modal features and intra-modal features produced by the bilinear layers. | [C-index] Gene: 0.695, pathology images: 0.578, gene + images: 0.723 | [C-index] Proposed: 0.723 lasso-Cox 0.700 inter-modal features: 0.708, [26]: 0.694, concatenation: 0.703 |
| 25 | Braman et al. 2021 [36] | T1 and T2 MRI images, DNA, clinical features | 176 patients from TCGA-GBM/LGG and BraTs. (Public) | Survival prediction of brain glioma tumors | Operation Attention Tensor Fusion | Extended the fusion method in [26] to four modalities and the orthogonal loss was added to encourage the learning of complementary uni-modal features. | [C-index] Radiology: 0.718, Pathology: 0.715, gene: 0.716, clinical: 0.702, Path+Clin: 0.690, all: 0.785 | [C-index] Proposed: 0.785, pathomic fusion [26]: 0.775, concatenation: 0.76 |
| 26 | Cao et al. 2021 [45] | fMRI images, clinical features | 871 subjects from ABIDE. (Public) | Classification of ASD and health controls. | Graph | Nodes features were composed of image features, while the edge weights were calculated by images and non-image features. | [Accuracy] Sites + gender + age + FI Q: 0.7456, sites + age + FIQ: 0.7534, sites + age: 0.7520 | [Accuracy] Proposed: 0.737, Parisot et al [44]: 0.704 |
| 27 | Parisot et al. 2018 [44] | fMRI images, clinical features | 1) 871 subjects from ABIDE. 2) 675 subjects from ANDI. (Public) | Classification of ASD and health control. Prediction of conversion to AD. | Graph | Nodes features were composed of image features, while the edge weights were calculated by images and non-image features. | [AUC] Image+sex+APOE4: 0.89, image+sex+APOE4+age: 0.85 (ADNI dataset) | [AUC] Proposed: 0.89, GCN: 0.85, MLP (Concatenation): 0.74 (ADNI dataset) |
| 28 | Chen et al. 2021 [27] | H&E images, genomic data | 1) - 4) 437, 1,022, 1,011, 515 and 538 patients from TCGA-BLCA, BRCA,GBMLGG ,LUAD and UCEC respectively (Public) | Survival prediction of five kinds of tumors. | Operation Attention | Co-attention mapping between WSIs and genomic features | [C-Index] Gene (SNN): 0.527, Pathology images: 0.614, All: 0.653 (overall prediction of five tumors) | [C-index] Proposed: 0.653, concatenation: 0.634, bilinear pooling: 0.621. (Overall prediction of five tumors) |
| 29 | Zhou et al. 2019 [63] | PET images, MRI images, SNP | 805 patients from ADNI （360 with complete multi-modalities）. (Public) | Classification of AD and its prodromal status | Operation | Learned features of every two modalities and all three modalities were concatenated at the 1st and 2nd fusion stage separately. | [Accuracy] MRI+PET+SNP > MRI+PET > MRI > MRI+SNP > PET+SNP > PET > SNP (Four-class classification) | [Accuracy] Proposed > MKL [88] > SAE [89] (Direct concatenation of learned uni-modal features) |
| 30 | Huang et al. 2020 [39] | CT images, clinical features, and lab test | 1,837 studies from 1,794 patients | Classification of the presence pulmonary | Operation | Compared 7 kinds of fusion, including early, intermediate and late fusion. Late elastic fusion performed the best. | [AUC] Images: 0.833, clinical and lab test: 0.921, all: 0.962. | [AUC] Early fusion: 0.899, late fusion: 0.947, joint fusion: 0.893. |
| | | results | | embolism | | | | |
| 31 | Lu et al. 2022 [73] | Pathology images, genomics data | 736 patients from TCGA-GBM/LGG (Public) | Survival prediction and grade classification of glioma tumors. | Operation Attention | Proposed a multi-modal transformer encoder for co-attention fusion. | [C-index] Images: 0.7385, gene: 0.7979, images+gene: 0.8266 (Same trend for classification task) | [C-index] Proposed: 0.8266 pathomic fusion [26]: 0.7994 |
| 32 | Cai et al. 2022 [51] | Camera/ dermatoscopic images, clinical features. | 1) 10,015 cases from ISIC2018, 2) 760 cases from a private dataset. (Partially Public) | Classification of skin wounds | Operation Attention | Two multi-head cross attention to interactively fuse information from images and metadata. | [AUC] Images: 0.944 clinical features: 0.964 images+clinical: 0.974 (Private dataset) | [AUC] Poposed: 0.974, metaBlock [90]: 0.968, concatenation: 0.964 (Private dataset) |
| 33 | Jacenkow et al.2022 [77] | X-ray images, full-text reports | 210,538 cases from MIMIC-CXR | Classification chest diseases | Attention | Unimodally pre-trained BERT model and fine-tune it to a multi-modal task. | [ACC] Images: 86 text: 85.1 images+text: 87.7 | [ACC] Proposed: 87.7, attentive [91]: 86.8 |
| 34 | Li et al. 2020 [60] | X-ray images, full-text reports | 1) 222,713 cases from MIMIC-CXR, 2) 3684 cases from Open I. (Public) | Classification of chest diseases | Attention | Used different pre-trained visual-text transformer | [AUC] Text: 0.974, VisualBert: 0.987 (MIMIC-CXR) | [AUC] VisualBERT [92], [93]: 0.987, LXMERT [94]: 0.984, UNITER : 0.985,PixelBERT:0.953(MIMIC-CXR) |

* No significant difference: p-value ≥ 0.05, significant difference, p-value < 0.05.

## 5. DISCUSSION AND FUTURE WORK

In the above sections, we reviewed recent studies using deep learning-based methods to fuse image and non-image modalities for disease prognosis and diagnosis. The feature-level fusion methods were categorized into operation-based, subspace-based, attention-based, tensor-based, and graph-based methods. The operation-based methods are intuitive and effective, but they might yield inferior performance when learning from complicated interactions of different modalities' features. However, such approaches (e.g., concatenation) are still used to benchmark new fusion methods. Tensor-based methods represent a more explicit manner of fusing multi-modal features, yet with an increased risk of overfitting. Attention-based methods not only fuse the multi-modal features but compute the importance of inter- and intra-modal features. Subspace-based methods tend to learn a common space for different modalities. The current graph-based methods employ graph representation to aggregate the features by incorporating prior knowledge in building the graph structure. Note that these fusion methods are not exclusive to each other, since some studies combined multiple kinds of fusion methods to optimize the prediction results. Compared with decision-level fusion for the decision-level fusion, feature-level fusion may gain benefits from the interaction between multi-modal features, while the decision-level fusion is more flexible for the combination of multi-modalities and thus robust to modality missing problems.

Although different fusion methods have different characteristics, how to select the optimal fusion strategy is still an open question in practice. There is no clue that a fusion method always performance the best. Currently, it is difficult to compare the performance of different fusion methods directly, since different studies were typically done on different datasets with different settings. Moreover, most of the prior studies did not use multiple datasets or external testing sets for evaluation. Therefore, more fair and comparative studies and benchmark datasets should be encouraged for multi-modal learning in the medical field. Furthermore, the optimal fusion method might be task/data dependent. For example, the decision-level fusion might be more suitable for multi-modality with less correlation. However, the theorical analysis and evaluation metrics are not extensively researched. Some study shows that the fusion at different layers or levels can significantly influence the results [61] [15] [39] [63]. The neural architecture search provides an option to automatically optimize the network structure. It has been applied in the image and non-image fusion in other fields [95] [96] [97], but it is under explored in multi-modal medical applications.

The reviewed studies showed that the performance of multi-modal models typically surpassed the uni-modal counterparts in the downstream tasks such as disease diagnosis or prognosis. On the other hand, some studies also mentioned that the model that fused more modalities may not always perform better than the ones with fewer modalities. In other words, the fusion of some modalities may have no influence or negative influence on multi-modal models [23], [18], [36] [45], [44]. It might be because the additional information introduces bias for some tasks. For example, Lu et al. [23] used the data of both primary and metastatic tumors for training to increase the top-k accuracy of the classification of metastatic tumors effectively. However, the accuracy decreased by 4.6% when biopsy site, a clinical feature, was added. Parisot et al. [44] and Cao et al. [45] demonstrated that the fusion of redundant information or data with noise (e.g., age, full intelligence quotient) led to defining inaccurate neighborhood systems of the population graph and further decreased the model performance. Meanwhile, additional modalities increase the network complexity with more trainable parameters, which may increase the training difficulties and the risk of overfitting. Braman et al. [36] used outer products to fuse uni-modal features. However, the outer products with three modalities yielded an inferior performance compared with the pairwise fusion and even uni-modal models. Thus, although multi-modal learning tends to benefit model performance, modality selection should consider the model capacity, data quality, specific tasks, etc. This is still an interesting problem worth more exploration.

A concern in this field is data availability. Although deep learning is powerful in extracting a pattern from complex data, it requires a large amount of training data to fit a reasonable model. However, data scarcity is always a challenge in the healthcare area, the situation of the multi-modal data is only worse. Over 50% of the reviewed studies used multi-modal datasets containing less than 1,000 patients. To improve the model performance and robustness with limited data, the pre-trained networks (e.g., image encoder Transformers and CNN networks pre-trained by natural images [98], text encoder BERT [57] and ClinicalBERT [59], and multi-modal encoder VisualBERT [92], LXMERT [93] and UNITER [94] pre-trained by natural image-visual pairs) were widely used by many studies instead of training from scratch with small datasets. Meanwhile, several studies [73] [47] [22], [23] deployed multi-task learning and showed improvements. Through sharing representations between related tasks, models generalized better on the original task. Also, many studies applied feature reduction and data augmentation techniques to avoid overfitting. To enlarge the paired dataset scale for multimodal fusion, combining multi-site data is a straightforward method, but the computational data harmonisation is worth to consider to eliminate the non-biological variances of multi-site data for a general and robust model [99]. Also. unsupervised methods [100] and semi-supervised methods [101] which have gained a great success in unimodal learning can also be applied to utilize the multi-modal data without labels. Meanwhile, transfer learning with larger datasets shared related knowledge can benefit the multi-modal learning. For example, Sharifi-Noghabi et al. [102] used multi-modal pan-drug data in training to enlarge the multi-modal dataset and achieve better performance of the response prediction in a drug-specific task than the model trained by drug-specific data only. Similarly, Cheerla et al. [25] noticed that the survival prediction of single cancers was improved by using all cancer data instead of using the single cancer data for training. Data missing is another problem of data availability. Complete datasets with all modalities available for every patient are not always guaranteed in routine practice. In the reviewed papers, random modality dropout [52] [37], multi-stage training [37] [36] [63], partial network training, data imputation [40], recurrent neural networks (RNN) [64], and autoencoders with reconstruction loss [72] [37] have been implemented to handle the missing data in multi-modal learning. However, the comparison of these methods and the influence of missing data in training and testing phases were not thoroughly investigated. Utilizing limited data effectively and efficiently is a practical but essential problem. This is a fast-growing direction attracting more and more attention.

Unimodal feature extraction is an essential prerequisite for fusion, especially for multi-modal heterogeneity. Proper preprocessing and feature extraction methods/networks are inevitable for the following fusion procedures. Both the standard feature extraction methods and learning-based feature extraction methods are commonly seen in deep fusion works. According to some reviewed works, different feature extraction methods can influence the fusion results significantly. For example, Cai et al. [51] observed that the ViT-based image encoder led to better fusion results than CNN-based encodes, and the fusion model using clinical features with soft one-hot encoding also outperformed hard encoding and word2vec. Li et al. [60] compared the contribution of different pre-trained language models to multi-modal fusion. The results showed that although the language model CliniclBERT by medical data outperformed the BERT model in unimodal prediction, it does not fit the pre-trained weights in the fusion stage and performed slightly worse in multi-modal prediction. To boost the fusion prediction, unimodal preprocessing and feature extraction should be carefully designed and evaluated. For better unimodal representation, different strategies can be considered. For example, the segmentation results tend to benefit the diagnostic feature extraction and ease the unimodal learning by providing regions of interests [103] [37] [62] [42]. The combination of conventional radiomics features and learning-based features achieved better performance [42] [62]. And the unimodal encoder trained by more data is more capable for representation generalization [63] [37] [101] [100]. Large unimodal datasets are always easier to obtain than multi-modal dataset. The leading techniques can be adapted to the unimodal feature extraction to boost the performance.

Explainability is another challenge in multi-modal diagnosis and prognosis. Lack of transparency is identified as one of the main barriers to deploying deep learning methods in clinical practice. An explainable model not only provides a trustworthy result but also helps the discovery of new biomarkers. In the reviewed papers, some explanation methods were used to show feature contributions to results. For image data, heatmaps generated with the class activation maps algorithm (CAM) were used to visualize the activated region of images that were most relevant to the models' outputs [55] [18] [26]. Li et al. [32] displayed the attention weights of patches to visualize the importance of every patch to the multi-instance learning of a WSI. The activated image region was compared with prior knowledge to see whether the models focused on the diagnostic characteristics of images. For non-image features, Holste et al. [15] used a permutation-based measure, Ghosal et al. [72] used learnable dropout layers, while Zhou et al. [104] and Chen et al. [26] implemented a gradient-based saliency method to get the score of feature importance. Especially, explainability in multi-modal learning helps explain and visualize the interaction between different modalities, Sappagh et al. [40], Cheerla et al. [52] and Braman et al. [36] displayed the importance of modalities with the performance of multi-modal models trained by different combination of modalities. Chen et al. [27] visualized the correlation of gene data and pathology image regions to reflect the known genotype-phenotype relationships of cancers. Although the usefulness of these explanations is still waiting to be validated in clinical practice, the development of more advanced meta-explanation through multi-modal information fusion can be a promising topic for future study as Yang et al. [105] mentioned in their medical explainable AI review.

# 6. CONCLUSION

This paper has surveyed the recent works of deep multi-modal fusion methods using the image and non-image data in medical diagnosis, prognosis, and treatment prediction. The multi-modal framework, multi-modal medical data, and corresponding feature extraction were introduced, and the deep fusion methods were categorized and reviewed. From the prior works, multi-modal data typically yielded superior performance as compared with the uni-modal data. Integrating multi-modal data with appropriate fusion methods could further improve the performance. On the other hand, there are still open questions to achieve a more generalizable and explainable model with limited and incomplete multi-modal medical data. In the future, multi-modal learning is expected to play an increasingly important role in precision medicine as a fully quantitative and trustworthy clinical decision support methodology.

# ACKNOWLEDGMENT

This work is supported by Leona M. and Harry B. Helmsley Charitable Trust grant G-1903-03793, NSF CAREER 1452485, and Veterans Affairs Merit Review grants I01BX004366 and I01CX002171.